\documentclass{asme07nmy}
\usepackage{amsmath}
\usepackage{epsfig} 
\usepackage{color}
\usepackage{multirow}
\usepackage{graphics}
\usepackage{graphicx}
\usepackage{subfigure}
\usepackage{cite}
\usepackage{slashbox}

\newcommand{\efi}{\end{figure}}
\newcommand{\be}{\begin{equation}}
\newcommand{\ee}{\end{equation}}
\newcommand{\bea}{\begin{eqnarray}}
\newcommand{\eea}{\end{eqnarray}}

\newcommand{\remove}[1]{}

\def\pmm{PM}
\def\pms{PMs}

\def\mompm{multi-operation-mode PM}
\def\mompms{multi-operation-mode PMs}
\def\Mompms{Multi-operation-mode PMs}

\conffullname{}
\confshortname{}
\confdate{}
\confmonth{}
\confyear{}
\confcity{}
\confstate{}
\confcountry{}

\papernum{}
\title{Kinematics, Workspace and Singularity Analysis of \\ a Multi-mode Parallel Robot}
\author{Damien Chablat 
\affiliation{
Laboratoire des Sciences du Num\'erique de Nantes\\
UMR CNRS 6004, 1 rue de la No\"e, 44321 Nantes \\
email: damien.chablat@cnrs.fr}
}
\author{Xianwen Kong
\affiliation{
School of Engineering and Physical Sciences \\ Heriot-Watt University\\
Edinburgh, UK,  EH14 4AS \\
email: X.Kong@hw.ac.uk}
}
\author{Chengwei Zhang\thanks{Currently at Dalian Huarui Heavy Industry Technical Centre, China}
\affiliation{
School of Engineering and Physical Sciences \\ Heriot-Watt University\\
Edinburgh, UK,  EH14 4AS\\
email: moonlight1780@163.com}
}

\begin{document}

\maketitle

\begin{abstract}
A family of reconfigurable parallel robots can change motion modes by passing through constraint singularities by locking and releasing some passive joints of the robot. This paper is about the kinematics, the workspace and singularity analysis of a 3-PRPiR parallel robot involving lockable Pi and R (revolute) joints. Here a Pi joint may act as a 1-DOF planar parallelogram if its lockable P (prismatic) joint is locked or a 2-DOF RR serial chain if its lockable P joint is released. 
The operation modes of the robot include a 3T operation modes to three 2T1R operation modes with two different directions of the rotation axis of the moving platform. The inverse kinematics and forward kinematics of the robot in each operation modes are dealt with in detail. The workspace analysis of the robot allow us to know the regions of the workspace that the robot can reach in each operation mode. A prototype built at Heriot-Watt University is used to illustrate the results of this work. 
\end{abstract}
\section{Introduction}
During the past two decades, reconfigurable Parallel Manipulators (PM) have received much attention from a number of researchers \cite{Ding}. Several classes of  \mompms{} have been proposed \cite{Fanghella,Herve,KongJMD,LiHerve,Gogu2010,Ruggiu2012,KongIDETC2011a,KongIDETC2012,KongJMR2015,Gan2013,Zeng2014,Fang2014}. Different names have been used for this class of PMs, such as PMs that change their group of motion, PMs with bifurcation of motion, or disassembly-free reconfigurable PMs. The main characteristics of \mompms{} include \cite{KongIDETC2011a,KongIDETC2012,KongJMR2015}: a) Fewer actuators are needed for the moving platform to realize several specified motion patterns; and b) Less time is needed in reconfiguring the PM since there is no need to disassemble the PM in the process of reconfiguration . 

Recent advances in the research on \mompms{} are mainly on the type synthesis and the reconfiguration analysis of \mompms{}, while a \pmm{} with two 1T1R operation modes has been used as a novel swivel head for machine tools \cite{Herve}.
For example, a general method for the type synthesis of \mompms{} has been proposed in \cite{KongJMD, KongIDETC2012, KongIDETC2011a}. \Mompms{} with a 3-DOF translational mode and a 3-DOF (degrees-of-freedom) planar mode or a 4-DOF 3T1R mode have been obtained. In \cite{LiHerve}, several \pms{} with two 3T1R (also Sch\"{o}nflies motion) operation mode which has three translational DOF and one rotational DOF) operation modes have been investigated. Several \pms{} with two 2T1R operation modes have been proposed in \cite{Gogu2010, Ruggiu2012}. In \cite{KongJMR2015}, 2-DOF 3-4R PMs with both spherical translation mode and sphere-on-sphere rolling mode were obtained. References \cite{Coste15, NurahmiMMT16} shows 4-DOF 3T1R \pms{} that  have an extra 2-DOF or 3-DOF motion mode in addition to the required 4-DOF 3T1R motion mode. It is noted the metamorphic \pms{} based on reconfigurable U (universal) or S (spherical) joints \cite{DaiMMT01, DaiMMT02} and the multi-mode \pms{} with lockable joints \cite{Callegari2014, KongJin2015} are in fact kinematically redundant \pms{} in nature and do not belong to \mompms{} that this paper focuses on. 
In the reconfiguration analysis, which refers to identifying all the operation modes and the transition configurations, of \pms{}, several methods have been proposed \cite{Coste15, NurahmiMMT16, KongMMT2014, Husty, Callegari2014, Caro2016, KongMMT2016, KongIDETC2016}. As pointed out in \cite{KongMMT2016, KongIDETC2016}, practical methods for switching a PM from one operation mode to another is still an open issue.

In \cite{CWZhang}, a \mompm{} was proposed, which is a revised version of the \mompm{} developed at Heriot-Watt University as shown in Fig. 1 of \cite{KongJin2015}. This \mompm{} can switch from one operation mode to another securely by using reconfigurable planar parallelograms and lockable R (revolute) joints. 
This paper will perform a systematic study on the inverse kinematics, forward kinematics, workspace and singularity analysis of this \mompm{}. 
  
In Section \ref{DE}, the description of a 3-PRPiR PM is given. Section \ref{KA} deals with the inverse and direct kinematic problem. In Section \ref{SA}, the singularity analysis is presented. Section \ref{WA} presents the workspace analysis to find out the connected regions reachable for a given working mode in each operation mode. Finally, conclusions are drawn on the advantages of this robot. 

\section{Description of a 3-PRPiR parallel robot} \label{DE}
\begin{figure}
\begin{center}
\includegraphics[scale=.45]{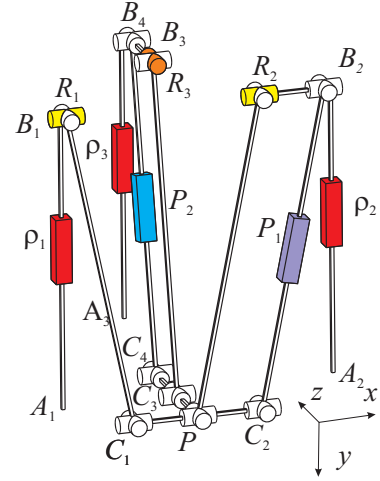}
\end{center}
\caption{A 3-RPiR multi-mode parallel robot with the three actuators in red, the passive joints in white and the lockable joints in other colors}
\label{Fig:Robot}
\end{figure}

The studied robot \cite{KongMMT2016} has three degrees-of-freedom and motion of the moving platform depends on the operation mode of the robot (Fig.~\ref{Fig:Robot}). The 3-RPiR\footnote{One of the Pi joint is replaced with an RR serial chain to simplify the structure without affecting the function of the robot.} multi-mode parallel robot is composed of a base and moving platform connected by three RPiR legs. Here R and Pi represent revolute and reconfigurable parallelogram joint respectively. The Pi joint may act as a 1-DOF planar parallelogram if its lockable P (prismatic) joint is locked or a 2-DOF RR serial chain if its lockable P joint is released. In total, there are three lockable revolute joints ($R_1$, $R_2$ and $R_3$) and two lockable prismatic joints ($P_1$ and $P_2$) which can be locked/released to change the operation modes of the robot. The change of operating mode is carried out only in the home pose shown in Fig.~\ref{Fig:Robot}. In the home pose, the axes of all the R joints, including those within the Pi joint, are parallel to two directions. 

The input joint variables are the three prismatic actuated joints $\rho_1$, $\rho_2$ and $\rho_3$  along the $y$-axis. Table \ref{Operation_Modes} describes the robot's mobility for four operation modes. 

\begin{table}%
\caption{Four operation modes of the multi-mode parallel robot.}
\begin{tabular}{|c|c|c|c|c|} \hline
\multicolumn{3}{|c|}{Operation modes} & \multicolumn{2}{c|}{Conditions} \\ \hline	
No & Translation   & Rotation  & Joints              &  Joints             \\
   &  along        & about     & locked              & released            \\ \hline
1  & $x$, $y$, $z$ & $--$      & $P_1$, $P_2$        & $R_1$, $R_2$, $R_3$ \\  \hline
2  & $x$, $y$      & $z$       & $R_1$, $R_2$, $P_2$ & $R_3$, $P_1$        \\  \hline
3  & $x$, $y$      & $x$       & $R_1$, $R_2$, $P_1$ & $R_3$, $P_2$        \\  \hline
4  & $y$, $z$      & $z$       & $R_3$, $P_2$        & $R_1$, $R_2$, $P_1$ \\  \hline
\end{tabular}
\label{Operation_Modes}
\end{table}

In a fixed reference frame, the positions of the fixed points $A_i$ ($i=1, 2, 3$) as well as the points $B_i$ controlled by the actuated prismatic joints are defined as 

\[\begin{array}{l}
{A_1} = \left[ { - {d_1}/2,0,0} \right]^T\\
{A_2} = \left[ {{d_1}/2 - {d_3},0,0} \right]^T\\
{A_3} = \left[ {0,0,{d_2} - {d_4}} \right]^T
\end{array}\]

\[\begin{array}{l}
{B_1} = \left[ { - {d_1}/2,{\rho _1},0} \right]^T\\
{B_2} = \left[ {{d_1}/2 - {d_3},{\rho _2},0} \right]^T\\
{B_3} = \left[ {0,{\rho _3},{d_2} - {d_4}} \right]^T
\end{array}\]

The motion of the end-effector $P$ depends on the operation modes of the robot. The coordinates of point $P$ will be given in details in the next subsections on the constraint equations for each operation mode.

\begin{figure}
\begin{center}
\includegraphics[scale=.45]{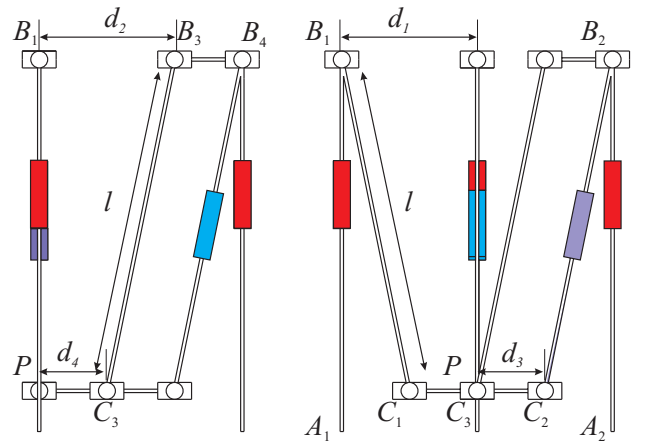}
\end{center}
\caption{Link parameters of the 3-RPiR multi-mode parallel robot.}
\label{Fig:Robot_parameters}
\end{figure}

In this paper, the following parameters (in normalized units) are used ${d_1} = 1/2$, ${d_2} = 1$, $l = 1$, ${d_3} = 1/10$ and ${d_4} = 1/10$ (Fig.~\ref{Fig:Robot_parameters}).
\subsection{Operaion Mode 1}
Operation mode 1 is a 3-DOF pure translation mode. That is to say the moving platform cannot rotate. This architecture is identical to the linear Delta robot found in the Renault Automation UraneSX \cite{Urane_SX} or the IRCCyN Orthoglide \cite {IEEE_Chablat,Chablat_Merlet}.

The coordinates of the points on the mobile platform are
\[\begin{array}{l}
C_1 = \left[ {x - {d_3},y,z} \right]^T\\
C_2 = \left[ {x + {d_3},y,z} \right]^T\\
C_3 = \left[ {x,y,z + {d_4}} \right]^T \\
P = \left[ {x,y,z} \right]^T
\end{array}\]
The constraint equations in operation mode 1 are obtained by calculating the distances between the points $B_i$ and $C_i$.
\[
||B_i - C_i || = l^2 \quad {\rm for ~i=1,2,3}
\]
i.e.
\bea
  \left\{
	\begin{array}{l}
{\left( {x + \frac{3}{{20}}} \right)^2} + {\left( {y - {\rho _1}} \right)^2} + {z^2} = 1\\
{\left( {x - \frac{3}{{20}}} \right)^2} + {\left( {y - {\rho _2}} \right)^2} + {z^2} = 1\\
{x^2} + {\left( {y - {\rho _3}} \right)^2} + {\left( {z - \frac{7}{{10}}} \right)^2} = 1 
	\end{array}
    \right.
\label{eq:c1}
\eea
\subsection{Operaion Mode 2}
For this operation mode, the moving platform undergoes a 3-DOF planar motion along the O-XY plane: it translates in the plane $(x,y)$  and rotate around the $z$-axis. We can write the position of $C_i$ and $P$ as
\[\begin{array}{l}
C_1 = \left[ {x - {d_3}\cos (\alpha ),y - {d_3}\sin (\alpha ),0} \right]^T\\
C_2 = \left[ {x + {d_3}\cos (\alpha ),y + {d_3}\sin (\alpha ),0} \right]^T\\
C_3 = \left[ {x,y,{d_4}} \right]^T \\
P     = \left[ {x,y,0} \right]^T
\end{array}\]
The constraint equations are
\bea
  \left\{
	\begin{array}{l}
{\left( {x - \frac{{\cos  \alpha}}{{10}} + \frac{1}{4}} \right)^2} + {\left( {y - \frac{{\sin  \alpha}}{{10}} - \rho 1} \right)_{}}^2 = 1\\
{\left( {x - \frac{3}{{20}}} \right)^2} + {\left( {y - {\rho _2}} \right)^2} = 1\\
{x^2} + {\left( {y - {\rho _3}} \right)^2} = \frac{9}{{25}} 
	\end{array}
    \right.
\label{eq:c2}
\eea
\subsection{Operaion Mode 3}
For this operation mode, the translations of the moving platform are in the plane $(x,y)$  and the rotation around the $x$-axis. Such motion is called PPR equivalent motion in \cite{Kong:2005}. We can write the position of $C_i$ and $P$ as
\[\begin{array}{l}
{C_1} = \left[ {x - {d_3},y,0} \right]^T\\
{C_2} = \left[ {x + d_3,y,0} \right]^T\\
{C_3} = \left[ {x,y + {d_4}\sin (\alpha ),{d_4}\cos (\alpha )} \right]^T \\
P     = \left[ {x,y,0} \right]^T
\end{array}\]
The constraint equations are
\bea
  \left\{
	\begin{array}{l}
{\left( {x + \frac{3}{{20}}} \right)^2} + {\left( {y - \rho 1} \right)^2} = 1\\
{\left( {x - \frac{3}{{20}}} \right)^2} + {\left( {y - \rho 2} \right)^2} = 1\\
{x^2} + {\left( {y +  \frac{{\sin  \alpha}}{{10}} - \rho 3} \right)^2} + {\left( {  \frac{{\cos  \alpha}}{{10}} - \frac{9}{{10}}} \right)^2} = 1 
\end{array}
    \right.
\label{eq:c3}
\eea

\subsection{Operaion Mode 4}
For operation mode 4, the moving platform undergoes a 3-DOF planar motion along the O-YZ plane. The translations of the moving platform are in the plane $(y,z)$  and the rotation around the $x$-axis. The position of $C_i$ and $P$ are
\[\begin{array}{l}
{C_1} = \left[ { - {d_3}\cos \alpha,y - {d_3}\sin \alpha,z} \right]^T\\
{C_2} = \left[ {{d_3}\cos \alpha,y + {d_3}\sin \alpha,z} \right]^T\\
{C_3} = \left[ {0,y,z + {d_4}} \right]^T \\
P     = \left[ {0,y,z} \right]^T
\end{array}\]
The constraint equations are
\bea
  \left\{
	\begin{array}{l}
{\left( { -  \frac{{\cos  \alpha}}{{10}} + 1/4} \right)^2} + {\left( {y -  \frac{{\sin  \alpha}}{{10}} - {\rho _1}} \right)^2} + {z^2} = \! 1\\
{\left( {y - {\rho _2}} \right)^2} + {z^2} = \!\!\frac{{391}}{{400}}\\
{\left( {y - {\rho _3}} \right)^2} + {\left( {z - 4/5} \right)^2} = \! 1
\end{array}
    \right.
\label{eq:c4}
\eea
\section{Inverse and direct kinematic analysis of the parallel robot in different operation modes} \label{KA}
\subsection{Inverse kinematic analysis}
For each of the four operation modes, there are $8$ solutions to the inverse kinematic model or working modes \cite{ICRA_1998}. From the constraint equations [Eqs.~(\ref{eq:c1})--(\ref{eq:c4})], it is quite straightforward to solve quadratic equations to find the inputs of the actuators. These solutions are given below without detailed derivation. 

\begin{figure}
\begin{center}
\includegraphics[scale=.45]{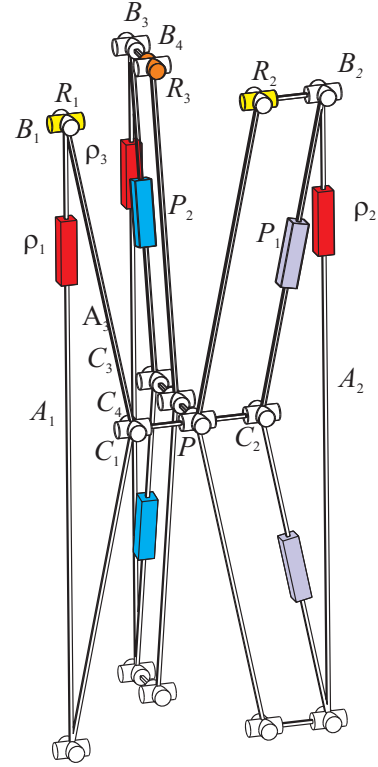}
\end{center}
\caption{The eight working modes of the robot associated with the home pose.}
\label{Fig:Working_mode}
\end{figure}

{\bf \noindent Operation Mode 1}\\
\[\begin{array}{l}
\rho_1=y \pm \sqrt {-400\,{x}^{2}-400\,{z}^{2}-120\,x+391} / 20  \\
\rho_2=y \pm \sqrt {-400\,{x}^{2}-400\,{z}^{2}+120\,x+391} /20  \\
\rho_3=y \pm \sqrt {-25\,{x}^{2}-25\,{z}^{2}+40\,z+9} / 5 
\end{array}\]
{\bf \noindent Operation Mode 2}\\
\[\begin{array}{l}
 \rho_1= y-\sin  \alpha/10 \pm  \\
 \!{\sqrt {\! 80 x \cos  \alpha \! + \! 4 \sin^{2}  \alpha \! \! - \! 400 {x}^{2} \! + \! 20 \cos  \alpha \! - \! 200 x \! + \! 371}}/{20}  \\
 \rho_2= y \pm \sqrt {-400\,{x}^{2}+120\,x+391}/20  \\
 \rho_3= y \pm \sqrt {-25\,{x}^{2}+9} /5 
\end{array}\]
{\bf \noindent Operation Mode 3}\\
\[\begin{array}{l}
\rho_1=y \pm \sqrt {-400\,{x}^{2}-120\,x+391} /20  \\
\rho_2=y \pm \sqrt {-400\,{x}^{2}+120\,x+391} /20  \\
\rho_3=y \pm \sin  \alpha /10  \\
+\sqrt {  \sin^{2}  \alpha  -100\,{x}^{2}+18\,\cos \alpha +18} /10 
\end{array}\]
{\bf \noindent Operation Mode 4}\\
\[\begin{array}{l}
\rho_1=y-\sin  \alpha /10  \\
\pm \sqrt {  \sin^{2}  \alpha   -100\,{z}^{2}+10\,\cos  \alpha +74} /10 \\
\rho_2=y \pm \sqrt {-100\,{z}^{2}+99} /10  \\
\rho_3=y \pm \sqrt {-400\,{z}^{2}+640\,z+119} /20  
\end{array}\]

Figure~\ref{Fig:Working_mode} shows the $8$ working modes of the robot for the home pose.

\subsection{Direct kinematic analysis}
Solving the constraint equations associated with each operation mode [Eqs.~(\ref{eq:c1})--(\ref{eq:c4})], one can find the locations of the moving platform for a given set of inputs.

For operation mode 1, the resolution of the direct kinematic analysis amounts to calculating the intersection of three spheres. An efficient geometric method is presented by Pashkevich in \cite{Robotica}. For the other operation modes, we obtain two values for the position according to $x$,$ $y or $z$. For the orientation of the moving platform, we obtain a quadratic equation as a function of the position parameters. Therefore, we usually have four real solutions for the orientation of the moving platform.

\begin{figure}
\begin{center}
\includegraphics[scale=.25]{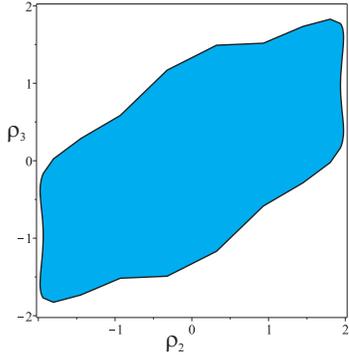}
\end{center}
\caption{Joint space of operation mode 1.}
\label{Fig:ART_1}
\end{figure}

\begin{figure}
\begin{center}
\includegraphics[scale=.25]{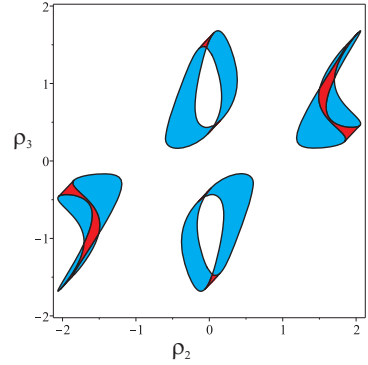}
\end{center}
\caption{Joint space of operation mode 2.}
\label{Fig:ART_2}
\end{figure}

\begin{figure}
\begin{center}
\includegraphics[scale=.25]{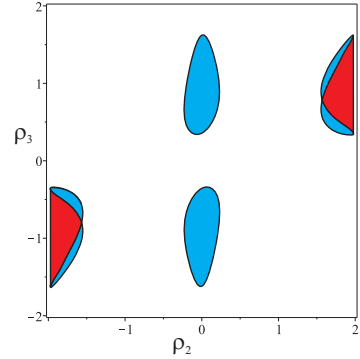}
\end{center}
\caption{Joint space of operation mode 3.}
\label{Fig:ART_3}
\end{figure}

\begin{figure}
\begin{center}
\includegraphics[scale=.25]{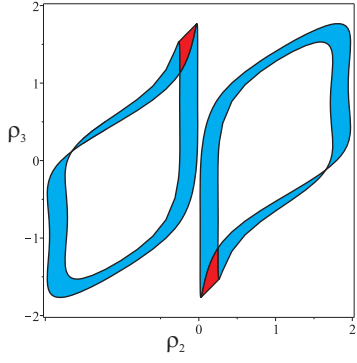}
\end{center}
\caption{Joint space of operation mode 4.}
\label{Fig:ART_5}
\end{figure}

Since the equations for all the four operation modes are too large to be displayed in an article and the solutions are well-documented in the literature, we will focus on the variation of the number of solutions to the direct kinematic model in the joint space. For clarity, we make a slice in the joint space by fixing $\rho_1 = 0$. This  is equivalent to the change of variable $\rho_2'=\rho_2-\rho_1$ and $\rho_3'=\rho_3-\rho_1$.
Figures \ref{Fig:ART_1}--\ref{Fig:ART_5} represent the joint space in blue, the regions where the direct kinematic model admits two real solutions and in red, the regions where it admits four real solutions.

In operation mode 1, there is only one region where the direct kinematic problem admits two solutions  (Fig.~\ref{Fig:ART_1}).

In operation mode 2, there are four connected regions where the direct kinematic problem admits either two or four solutions (Fig.~\ref{Fig:ART_2}). For two of them, there are one hole inside. This property can cause problems if the robot is to be protected by introducing limits on the active joints.

In operation mode 3, there are four connected regions  (Fig.~\ref{Fig:ART_3}). In two regions, the direct kinematic problem admits either two or four solutions and the other two regions, the direct kinematic problem admits only two solutions.

In operation mode 4, there are two connected regions with one hole  (Fig.~\ref{Fig:ART_5}). The direct kinematic problem admits either two or four solutions.

\begin{figure}[!ht]
\begin{center}
\includegraphics[scale=.5]{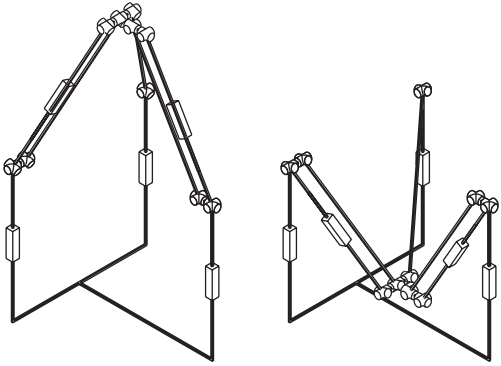}
\end{center}
\caption{Example of joint configuration with two direct kinematic solutions in operation mode 1.}
\label{Fig:Assembly_1}
\end{figure}

\begin{figure}[!ht]
\begin{center}
\includegraphics[scale=.5]{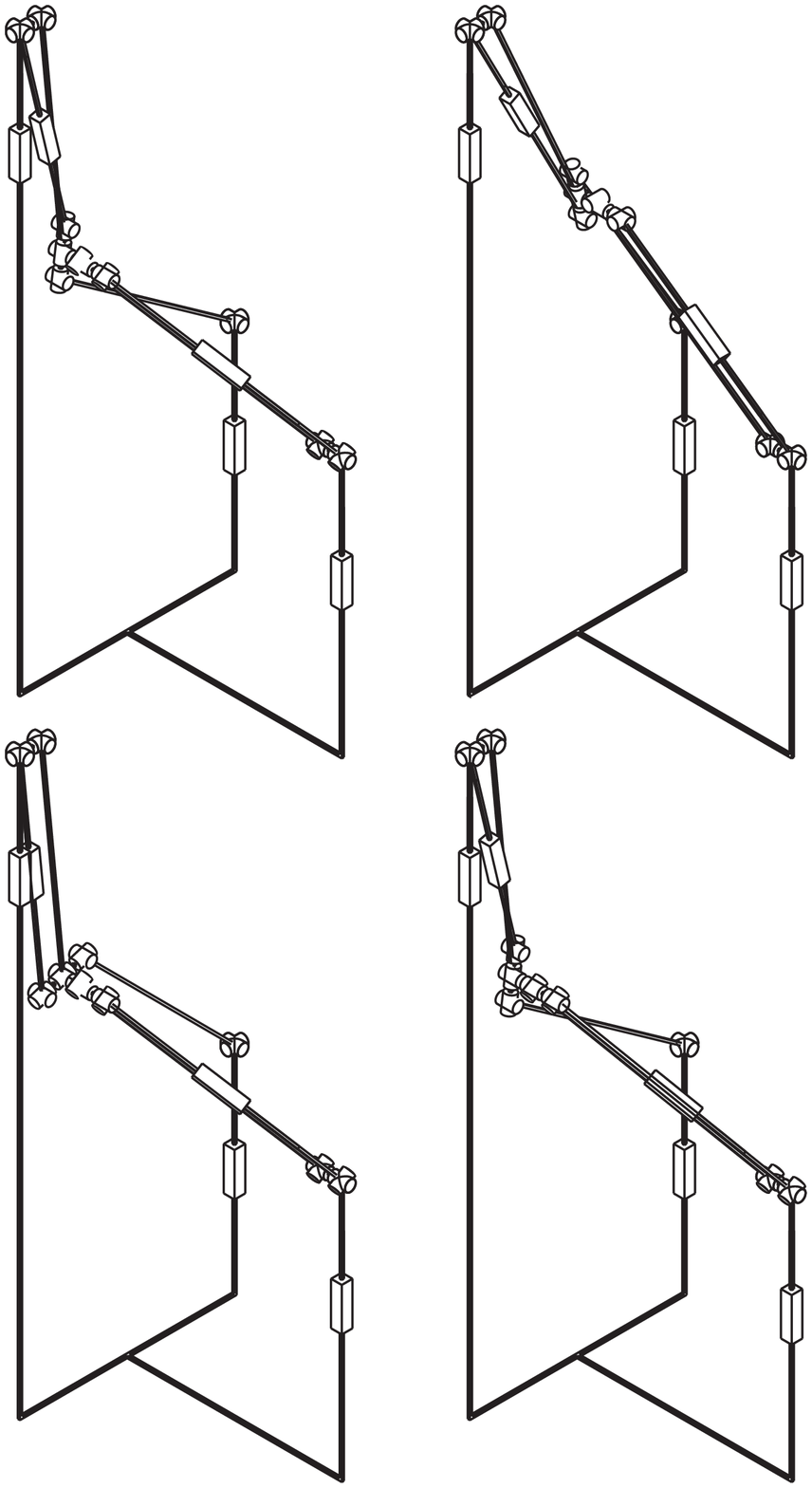}
\end{center}
\caption{Example of joint configuration with four direct kinematic solutions in operation mode 2.}
\label{Fig:Assembly_2}
\end{figure}

\begin{figure}[!ht]
\begin{center}
\includegraphics[scale=.5]{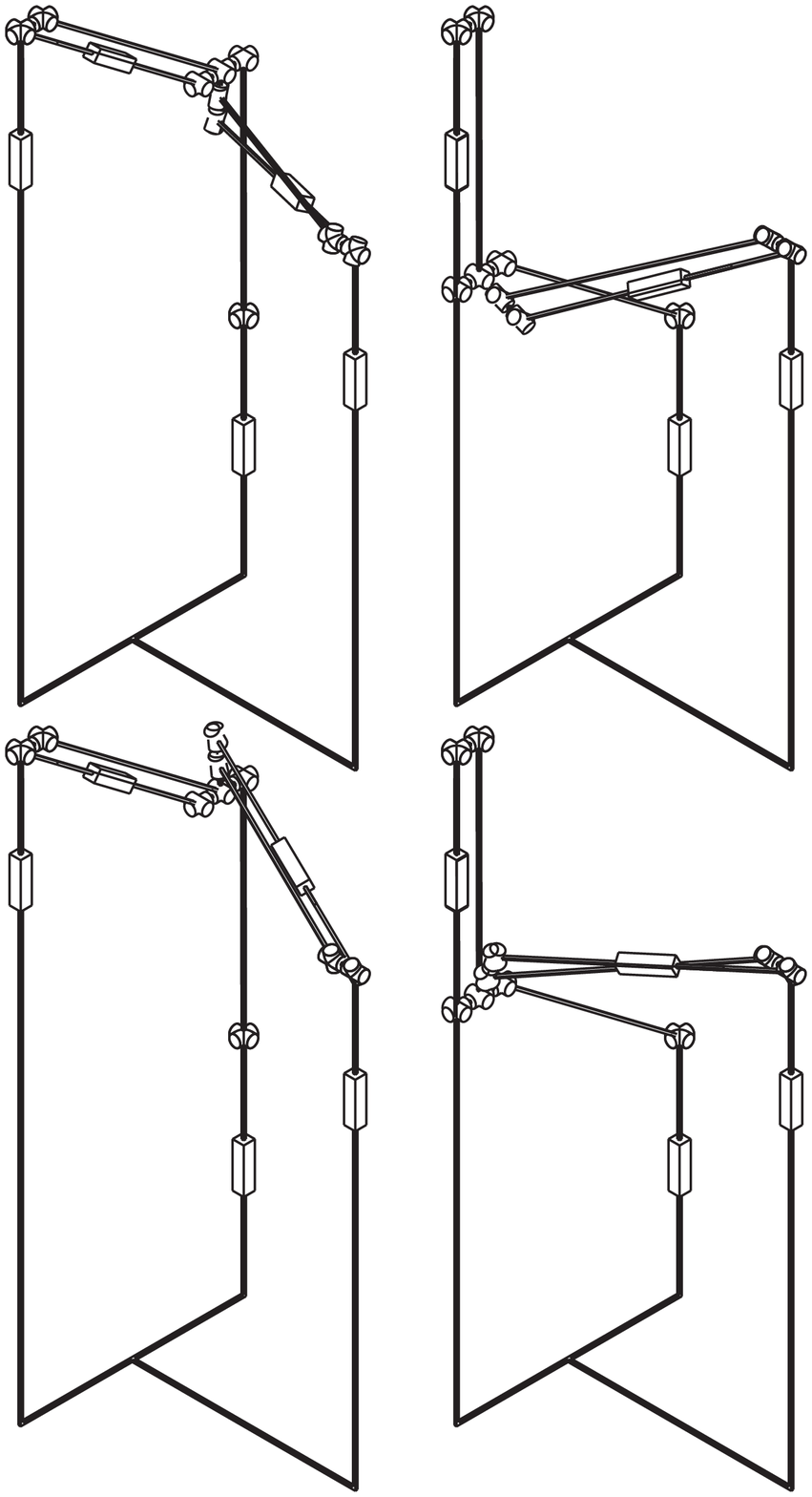}
\end{center}
\caption{Example of joint configuration with four direct kinematic solutions in operation mode 3.}
\label{Fig:Assembly_3}
\end{figure}

\begin{figure}[!ht]
\begin{center}
\includegraphics[scale=.5]{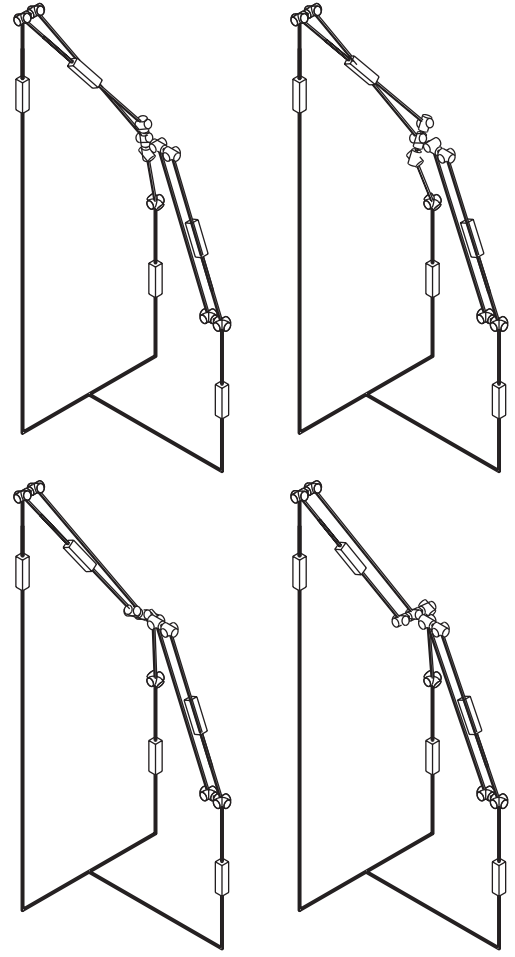}
\end{center}
\caption{Example of joint configuration with four direct kinematic solutions in operation mode 4.}
\label{Fig:Assembly_5}
\end{figure}

It is noted that except operation mode 1, the regions in the joint space are not connected. Figures~\ref{Fig:Assembly_1}--\ref{Fig:Assembly_5} shows several postures of the robot when we have the two or four solutions for the direct kinematics in these four operation modes. 

\section{Singularity analysis of the parallel robot in different operation modes} \label{SA}
The singularity analysis of conventional parallel robots has been well-documented in the literature. The singularity analysis of a multi-mode parallel robot involves the singularity analysis in each operation mode of the robot. 
In each operation mode,  we first find the parallel and serial Jacobin matrix, named $\bf A$ and $\bf B$ respectively \cite{Wenger:1997}, by differentiating the constraint equations [Eqs.~(\ref{eq:c1})--(\ref{eq:c4})] with respect to time. Then we obtain the parallel and serial singularities by studying the determinant of these matrices. In each operation mode, the serial singularities are located at the boundaries of the workspace while the parallel singularities divide it.

As the robots studied have several solutions to the direct and inverse kinematic analysis, the singularity conditions are defined in the cross product of the joint space and workspace. Their projections into the workspace are not requested because we cannot distinguish the curves associated with a given working mode.
\subsection{Operation mode 1}
The singular configurations in operation mode 1 of this robot are the same as the conventional parallel robot presented in \cite{Pashkevich:2015}. We have the serial singularities when the legs are orthogonal to the actuators and the parallel singularities when the legs are in the same plane (flat position) or all parallel (bar position).

{\bf \noindent Serial singularities} occur if and only if
\[\begin{array}{l}
{\rho_3} - y = 0{\rm,} \quad {\rho_2} - y = 0 \quad {\rm or} \quad {\rho_1} - y = 0
\end{array}\]

{\bf \noindent Parallel Singularities} occur if and only if
\[140  x{\rho_1} + 30  {\rho_1}  z - 140  x{\rho _2} + 30  {\rho _2}  z - 60  z{\rho _3} - 21  {\rho _1} - 21  {\rho _2} + 42  y = 0\]

In this operation mode, the determinant of the matrix $\bf A$ cannot be factorized. In a single equation, we have the singularity locus associated with the eight working modes.
\subsection{Operation mode 2}
{\bf Serial singularities} occur if and only if 
\[\begin{array}{l}
{\rho _3} - y = 0{\rm,}\;\;\; \quad {\rho _2} - y = 0 \quad {\rm or} \\
 - 10  y + \sin  \alpha + 10  {\rho _1} = 0
\end{array}\]
Only the serial singularities associated with the first leg $\rho_1$ depend on the orientation of the moving platform.

{\bf \noindent Parallel Singularities} occur if and only if 
\[\begin{array}{l}
20  x({\rho _2} - {\rho _3}) + 3  ({\rho _3} - 3 y) = 0 \quad {\rm or}\\
- 4  x \sin \alpha - 4  \cos \alpha{\rho _1} + 4  y\cos \alpha - \sin \alpha = 0
\end{array}\]
The determinant of the matrix $\bf A$ is factorized into two components. This properties means that the workspace is divided into at least  four regions. We  notice that the first component does not depend on the orientation of the moving platform.
\subsection{Operation mode 3}
{\bf Serial singularities} occur if and only if 
\[\begin{array}{l}
{\rho _2} - y = 0 {\rm ,} \quad
{\rho _1} - y = 0 \quad {\rm or}\\
 - 10  y - \sin  \alpha + 10  {\rho _3} = 0
\end{array}\]
As the axis of rotation of the moving platform is parallel to the $x$-axis, only the serial singularities related to the third leg depend on the orientation of the moving platform.

{\bf \noindent Parallel Singularities} occur if and only if 
\[\begin{array}{l}
20  {\rho _1}  x - 20  {\rho _2}  x - 3  {\rho _1} - 3  {\rho _2} + 6  y = 0 \quad {\rm or}\\
 - 10  \cos  \alpha{\rho _3} + 10  y\cos  \alpha + 9  \sin  \alpha = 0
\end{array}\]
\subsection{Operation mode 4}
{\bf Serial singularities}  occur if and only if 
\[\begin{array}{l}
{\rho _3} - y = 0 {\rm ,} \quad
{\rho _2} - y = 0 \quad {\rm or}\\
 - 10  y + \sin  \alpha + 10  {\rho _1} = 0
\end{array}\]
As the axis of rotation of the moving platform is parallel to the $z$-axis, only the serial singularities related to the first leg depend on the orientation of the moving platform.

{\bf \noindent Parallel Singularities}  occur if and only if 
\[\begin{array}{l}
5{\rho _{2}}z - 5 z{\rho _3} - 4 {\rho _2} + 4 y = 0 \quad {\rm or}\\
 - 4 \cos  \alpha{\rho _1} + 4  y\cos  \alpha - \sin  \alpha = 0
\end{array}\]
\section{Workspace analysis of the parallel robot in different operation modes} \label{WA}
The workspace analysis of a multi-mode parallel robot requires analyzing the workspace of the robot in each operation mode. In each operation mode, one needs to separate the postures of the robot according to the current working modes. We recall here the definition of the aspects $A_{ij}$ for parallel robots with several working modes $WM_i$ \cite{ICRA_1998}, which are defined as the maximal
sets in the product of the workspace, $W$, by the joint space, $Q$ so that
\begin{itemize}
\item ${\bf A}_{ij} \subset W \cdot Q$;
\item ${\bf A}_{ij}$ is connected;
\item ${\bf A}_{ij} = \left\{
                   ({\bf X}, {\bf q}) \in WM_i \setminus \det({\bf A}) \neq 0
                  \right\}$
\end{itemize}
where a {\em working mode}, noted $WM_i$, is the set of postures for which the sign of ${\bf B}_{jj}$ ($j = 1$ to $n$) does not change
and ${\bf B}_{jj}$ does not vanish. 

Since that it is impossible to change the working modes without disassembling this multi-mode parallel robot, we will study only the working modes depicted in Fig.~\ref{Fig:Robot} for simplicity reasons. With the SIROPA library, we can make an algebraic cylindrical decomposition (CAD) \cite{Cbook:75,Manubens} in the space which includes the joint space and the workspace taking into account the serial and parallel singularities. In this paper, we assume there is no limitation on the range of motion of the active and passive joints. Therefore, we can study the workspace without considering the translation along the $y$-axis.

In each cross-section of the workspace, we have represented the parallel aspects \cite{ICRA_1998}, i\@.e\@., the largest regions without singularity in the Cartesian space. The boundaries of these regions are the parallel singularities. Curves may exist that do not divide the workspace. These curves are associated with the other operation modes. These curves are obtained by the discriminant varieties of the constraint equations with the serial and parallel singularities \cite{Lazart:2007}.

\begin{figure}[!ht]
\begin{center}
\includegraphics[scale=.25]{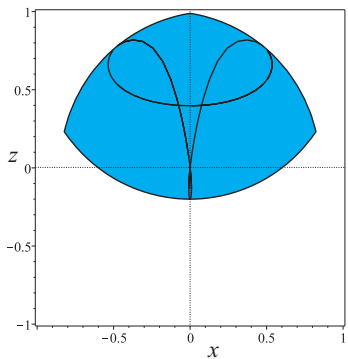}
\end{center}
\caption{Workspace in operation mode 1.}
\label{Fig:CART_1}
\end{figure}

In operation mode 1, the workspace is a single region (Fig.~\ref{Fig:CART_1}). Parallel singularities exist only when one leg is towards the positive $y$ and the other two towards the negative $y$ (and vice versa). These correspond to the curves in the center of the figure.

\begin{figure}[!ht]
\begin{center}
\includegraphics[scale=.25]{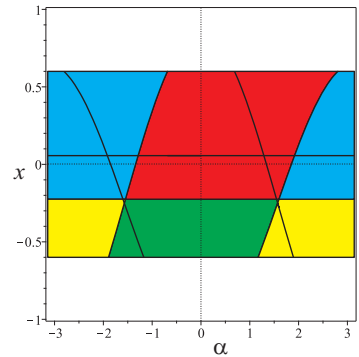}
\end{center}
\caption{Workspace in operation mode 2.}
\label{Fig:CART_2}
\end{figure}

In operation mode 2, the workspace is divided in four regions (Fig.~\ref{Fig:CART_2}). Please note the continuity of certain regions due to the cyclicity of the angles.

\begin{figure}[!ht]
\begin{center}
\includegraphics[scale=.25]{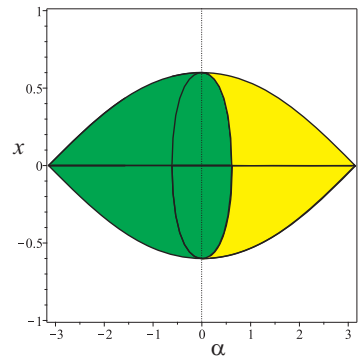}
\end{center}
\caption{Workspace in operation mode 3.}
\label{Fig:CART_3}
\end{figure}

The workspace in operation mode 3 consists of only two regions (Fig.~\ref{Fig:CART_3}). For $x = 0$, the moving platform can rotate mainly in the negative direction.

\begin{figure}[!ht]
\begin{center}
\includegraphics[scale=.25]{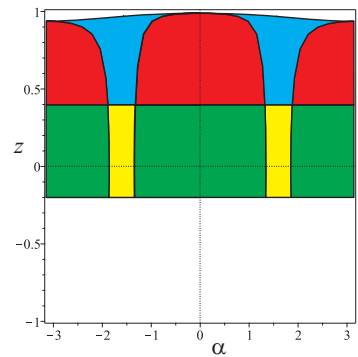}
\end{center}
\caption{Workspace in operation mode 4.}
\label{Fig:CART_5}
\end{figure}

The workspace in operation mode 4 has six aspects (Fig.~\ref{Fig:CART_5}). These aspects cannot be distinguished by the signs of the two components of the parallel singularities. This is the reason why we have two regions in blue, yellow, red and green each.

\begin{figure}[!ht]
\begin{center}
\includegraphics[width=9cm]{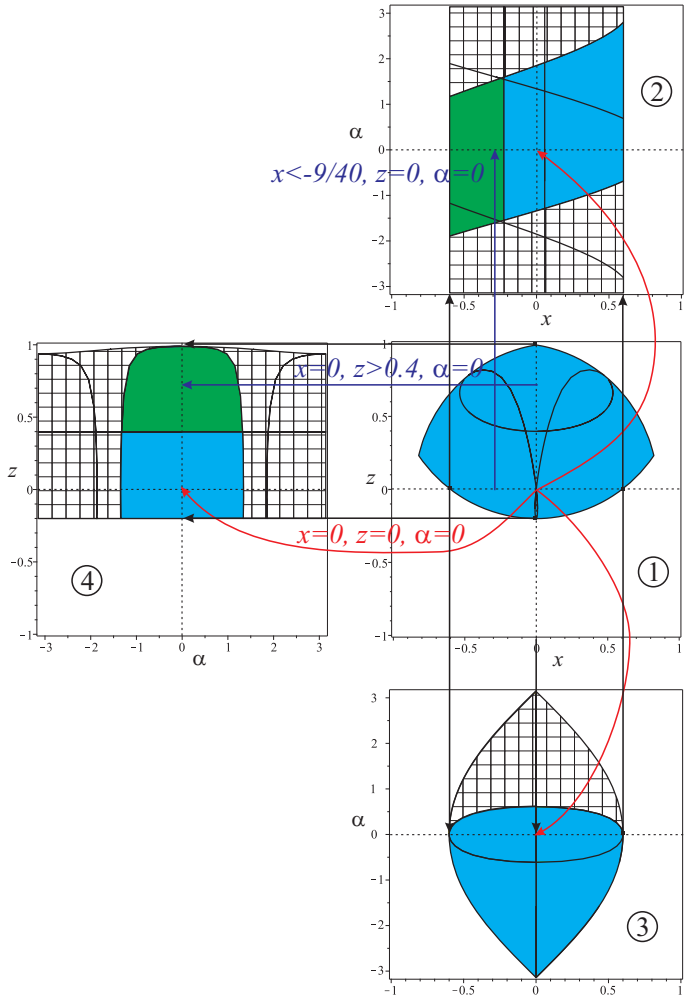}
\end{center}
\caption{Transition among the four operation modes.}
\label{Fig:Connect}
\end{figure}

The mult-mode parallel robot can switch among its four operation modes in its home pose in which $x = 0$, $z = 0$ and $\alpha=0$. In Fig.~\ref{Fig:Connect}, starting from the home pose, the red arrows show  the corresponding postures for each operation mode. Regions with grids are the regions of the workspace that the robot cannot reach without crossing a parallel singularity.

Starting from the configurations with $z = 0$ in operation mode 1, it is possible to switch to operation modes 2 and 3. 
It is noted for any values of $x$, only one aspect in operation mode 3 is reachable. 
For operation mode 2, two aspects, one with $x<-9/40$ and another with $x>-9/40$, are reachable.

Starting from the configurations with $x = 0$ in operation mode 1, it is possible to switch to operation mode 4. The robot can reach two aspects in operation mode 4: the green one with $z> (528 / 35 - 8 \sqrt{165} / 7)$ and the blue with $z< (528 / 35 - 8 \sqrt{165} / 7)$.

Based on the above results, we conclude that it is possible to change the aspects for a given operation mode by passing through operation mode 1. This property increases the workspace of the robot theoretically. However, this is not practical because the robot will have to change operation modes twice. We notice that the ranges of translation in operation modes 2, 3 and 4 are within the ranges defined by the intersections of the workspace in operation mode 1 with the $x$- and $z$- axes.

\section{Conclusions} \label{CO}
In this paper, we have presented the kinematics of a multi-mode parallel robot that can change operation modes by using reconfigurable Pi joints and lockable R joints. Changes in operation modes are realized by locking/releasing certain lockable joints. We have investigated the singularities of the parallel robot in different operation modes from the study of the determinant of the Jacobin matrices. For an operation mode  resulting from the ``home" configuration of the robot, we have decomposed the workspace into aspects and  represented their projection in the workspace. 
The conditions for switching from one operation mode to another have been identified. The aspects that the robot can achieve in a given operation mode without going through a parallel singularity have also been obtained.

This work, together with the literature on the reconfiguration analysis of multi-mode parallel mechanisms \cite{NurahmiMMT16, KongMMT2014, Husty, Caro2016, KongMMT2016, KongIDETC2016}, provide a comprehensive framework for the analysis of multi-mode parallel robots. 
\section*{Acknowledgments}
XK would like to thank the Engineering and Physical Sciences Research Council (EPSRC), United Kingdom, for the support under grant No.EP/K018345/1.

\end{document}